\documentclass[10pt,twocolumn,letterpaper]{article}

\usepackage{iccv}              % To produce the CAMERA-READY version

\usepackage{booktabs} % for better looking tables
\usepackage{multirow} % for multi-row cells
\usepackage{bbm}
\usepackage{marvosym}

\usepackage{appendix}

\newcommand\blfootnote[1]{%
  \begingroup
  \renewcommand\thefootnote{}\footnote{#1}%
  \addtocounter{footnote}{-1}%
  \endgroup
}
\definecolor{iccvblue}{rgb}{0.21,0.49,0.74}
\usepackage[pagebackref,breaklinks,colorlinks,allcolors=iccvblue]{hyperref}

\def\paperID{5149} % *** Enter the Paper ID here
\def\confName{ICCV}
\def\confYear{2025}

\title{Scale Your Instructions: Enhance the Instruction-Following Fidelity  of Unified Image Generation Model by Self-Adaptive Attention Scaling}

\author{ Chao Zhou\textsuperscript{\rm 1}, Tianyi Wei\textsuperscript{\rm 2,}\textsuperscript{\Letter}, Nenghai Yu\textsuperscript{\rm 1} \\
	\normalsize\textsuperscript{\rm 1}University of Science and Technology of China  \ \normalsize\textsuperscript{\rm 2}Nanyang Technological University  \  \\
	{\tt\small\{chaozhou@mail., ynh@\}ustc.edu.cn }, {\tt\small tianyi.wei@ntu.edu.sg} \\
}

\graphicspath{{fig/}}

\begin{document}

\twocolumn[{
	\renewcommand\twocolumn[1][]{#1}
	\maketitle
	\vspace{-2em}
	\setlength\tabcolsep{0.5pt}
	\centering
	\small
	\begin{tabular}{c}
        \includegraphics[width=\linewidth]{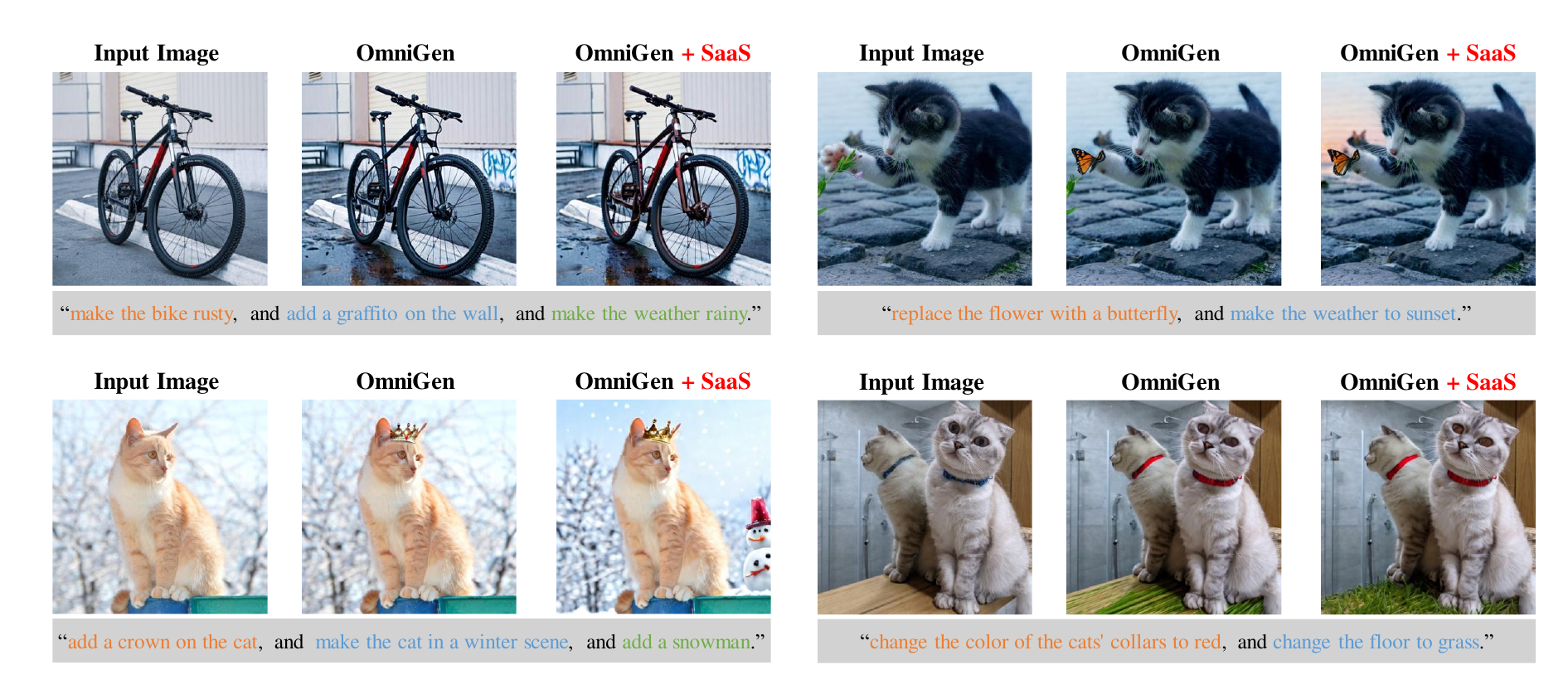}
        \end{tabular}
	\vspace{-1.5em}
	\captionof{figure}{\textbf{Results of SaaS.} OmniGen \cite{xiao2024omnigen} has a tendency to overlook some sub-instructions, while our SaaS can effectively mitigate this issue. Notably, SaaS does not require any extra training or test-time optimization. Zoom in for better visualization. }
        \vspace{1.5em}
	\label{fig:SaaS_results}
}]

\maketitle

\blfootnote{\Letter~Tianyi Wei is the corresponding author.}

\begin{abstract}

Recent advancements in unified image generation models, such as OmniGen, have enabled the handling of diverse image generation and editing tasks within a single framework, accepting multimodal, interleaved texts and images in free form. This unified architecture eliminates the need for text encoders, greatly reducing model complexity and standardizing various image generation and editing tasks, making it more user-friendly.
However, we found that it suffers from text instruction neglect, especially when the text instruction contains multiple sub-instructions.
To explore this issue, we performed a perturbation analysis on the input to identify critical steps and layers. By examining the cross-attention maps of these key steps, we observed significant conflicts between neglected sub-instructions and the activations of the input image. 
In response, we propose \textbf{Self-Adaptive Attention Scaling (SaaS)}, a method that leverages the consistency of cross-attention between adjacent timesteps to dynamically scale the attention activation for each sub-instruction. Our SaaS enhances instruction-following fidelity without requiring additional training or test-time optimization. 
Experimental results on instruction-based image editing and visual conditional image generation validate the effectiveness of our SaaS, showing superior instruction-following fidelity over existing methods. The code is available at ~\href{here}{https://github.com/zhouchao-ops/SaaS.}

\end{abstract}    
\section{Introduction}
\label{sec:intro}

\begin{figure*}[h]
    \centering
    \includegraphics[width=\linewidth]{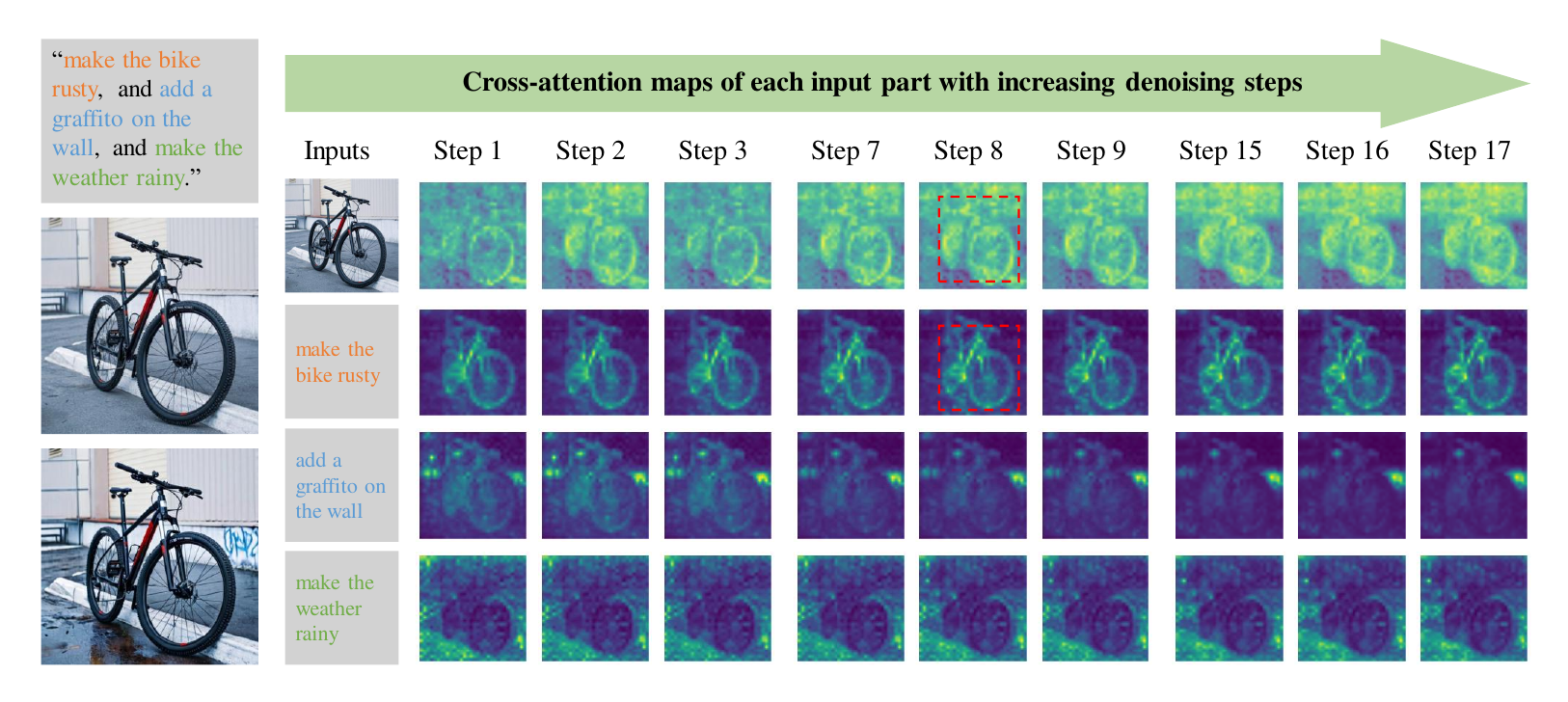}
    \caption{\textbf{Cross-attention maps for the input image and different sub-instructions.} We can get three key observations: (a) we can pre-identify the regions where each sub-instruction will appear according to the corresponding cross-attention map; (b) the regions of activation for the neglected sub-instruction are highly conflicting with those for the input image, where the input image dominates (red box); (c) the cross-attention maps remain highly consistent across adjacent timesteps.}
    \label{fig:crossattn}
    \vspace{-1em}
\end{figure*}

In recent years, image generation models have advanced rapidly. Using the Latent Diffusion Model (LDM) series \cite{rombach2021highresolution,podell2023sdxlimprovinglatentdiffusion,esser2024scalingrectifiedflowtransformers} as a benchmark, researchers have continuously improved the generated image quality. However, this progress has come at the cost of increasing model size and a growing reliance on larger, more complex text encoders \cite{radford2021learningtransferablevisualmodels,2020t5} to process text instructions. Moreover, for complex downstream tasks such as image editing and visual conditional image generation, these models often require additional structures \cite{zhang2023adding, ruiz2022dreambooth} or specialized methods \cite{Tumanyan_2023_CVPR, cao_2023_masactrl, hertz2022prompttopromptimageeditingcross}, making them less accessible and user-friendly.

Unlike the LDM series, unified image generation models such as OmniGen \cite{xiao2024omnigen} are trained on large unified output datasets, enabling them to handle diverse and complex downstream tasks within a single diffusion framework. Notably, OmniGen achieves this with remarkable efficiency, featuring a minimalistic yet powerful architecture composed of only two core components: a VAE \cite{pandey2022diffusevae} and a transformer model, without relying on additional text encoders. This streamlined architecture allows OmniGen to accept interwoven text prompts and image inputs as conditions for guiding image generation. Achieving comparable generation quality, OmniGen balances a lightweight design with enhanced user-friendliness.

% However, as illustrated in Fig. \ref{fig:SaaS_results}, Omnigen tends to overlook certain sub-instructions.  Through our analysis of Omnigen, we identified that this issue arises from conflicts between the activated regions of the input image tokens and the sub-instruction tokens on cross-attention maps.
As an all-in-one editing model, OmniGen demonstrates strong instruction-based image editing capabilities. However, as shown in Fig. \ref{fig:SaaS_results},  it frequently overlooks specific text instructions, particularly when handling multiple sub-instructions within a single prompt. To uncover the root causes of this issue, we conducted input perturbation experiments to pinpoint critical steps and layers in the denoising process. By further analyzing cross-attention maps at these key stages, we examined how generated pixels correlate with different input tokens, shedding light on the underlying mechanisms behind instruction adherence and omission.

Interestingly, our investigation revealed that the tendency to overlook instructions arises from significant conflicts between the activated regions on the cross-attention maps for the neglected sub-instructions and the input image. As illustrated in Fig. \ref{fig:crossattn}, the brightness of the maps reflects the magnitude of the activation values, with brighter regions indicating higher activations.  In the red-boxed area (the bike region of the generated image), the input image exhibits much stronger activations than the neglected sub-instruction, effectively suppressing its influence.  Additionally, we made two key observations: first, the regions with high activation values correspond roughly to areas where the sub-instructions influence the generated image; second, there is notable consistency in cross-attention between adjacent timesteps.

To address the issue of neglected sub-instructions, we propose \textbf{Self-Adaptive Attention Scaling (SaaS)}, a method that enhances the instruction-following fidelity of unified image generation models like OmniGen without requiring additional training or test-time optimization.
Building on the previously observed conflicts between the activation regions of text instructions and input images in the cross-attention maps, we adaptively scale the cross-attention values corresponding to the instructions during the denoising process.
This approach is essentially a free lunch for inference-time scaling, as it leverages the consistency of the cross-attention maps between adjacent denoising timesteps. At timestep $t$, we extract the mask for each sub-instruction and calculate the scaling factor. At timestep $t-1$, we apply the scaling factor to the activation values within the masked region of the corresponding sub-instruction. Masks scaling factors are iteratively updated throughout the denoising process.

Experimental results demonstrate that SaaS significantly enhances instruction-following fidelity across both image editing and visual conditional image generation tasks, ensuring more precise and consistent outputs.

Our contributions can be summarized as follows.
\begin{itemize}
    \item We identified for the first time that unified image generation models like OmniGen tend to overlook text instructions and confirmed the vital steps and layers in the denoising process through input perturbation analysis. 
    \item  We attributed the tendency to overlook instructions to conflicts between the activated regions of the neglected sub-instructions and the input image in the cross-attention maps, as revealed by analyzing the cross-attention maps of vital steps and layers.
    \item We propose SaaS, a novel self-adaptive attention scaling method to enhance instruction-following fidelity without any additional training or test-time optimization. 
    \item Qualitative and quantitative results demonstrate the effectiveness of the proposed SaaS.
    % \item We conducted extensive experiments on instruction-based image editing and visual conditional image generation, and both qualitative and quantitative results demonstrate the effectiveness of our SaaS.
\end{itemize}

\section{Related Work}
\label{sec:Related Work}

\noindent\textbf{Unified Image Generation.}
Unified input has long been a key goal in image generation. Early methods like T2I-Adapter \cite{mou2023t2iadapterlearningadaptersdig} and ControlNet \cite{zhang2023adding} rely on additional structures, while inversion methods \cite{mokady2022nulltextinversioneditingreal, rout2024rfinversion, hubermanspiegelglas2024editfriendlyddpmnoise} embed input image information by finding a suitable starting point in diffusion.
After years of development \cite{esser2020taming,oord2018neuraldiscreterepresentationlearning,wei2023hairclipv2,chang2022maskgitmaskedgenerativeimage,sun2024generativemultimodalmodelsincontext,brooks2022instructpix2pix}, unified image generation models \cite{xie2024showo,xiao2024omnigen,wang2024emu3,Chameleon_Team_Chameleon_Mixed-Modal_Early-Fusion_2024,Zhou2024TransfusionPT} have demonstrated significant potential. These models typically tokenize text and image inputs to form a unified sequence. Specifically, Emu3 \cite{wang2024emu3} generates both text and image autoregressively, while Show-o \cite{xie2024showo} generates text autoregressively and images via discrete diffusion separately. 
OmniGen \cite{xiao2024omnigen} focuses on the field of image generation, utilizing a flow-matching diffusion method \cite{lipman2023flowmatchinggenerativemodeling} for high-quality image output. Compared with those models that unify text and image generation, OmniGen demonstrates stronger image generation capabilities with faster processing speeds. In this paper, we focus on enhancing OmniGen’s instruction-following fidelity and investigate underlying challenges in this new unified image generation framework.

\noindent\textbf{Efforts to Instruction-following Fidelity.}
In diffusion models, refining attention maps has been shown to enhance instruction-following fidelity to some extent~\cite{wei2024enhancing,wei2025freeflux,yanmingRoCCRobustCovert2024}. Guo et al. \cite{guo2023focusinstructionfinegrainedmultiinstruction} by adjusting cross-attention, ensuring that instructions are properly aligned with relevant image regions. 
In Visual-Language Models \cite{abdin2024phi3technicalreporthighly,chen2024far,Qwen2-VL,Qwen2.5-VL,chen2024internvl}, some visual tokens are redundant~\cite{kaduri2024_vision_of_vlms,10.1145/3664647.3681712}, and reducing token redundancy can achieve better instruction-following fidelity. In particular, Yang et al. \cite{yang2024enhancinginstructionfollowingcapabilityvisuallanguage} 
address this by condensing redundant image tokens, directing the model’s focus to key visual features, and thereby improving fidelity. However, these approaches are model-specific and not directly applicable to OmniGen. 
In this paper, we bridge the gap in the instruction-following fidelity of the latest unified image generation models represented by OmniGen.

\section{Method}

\subsection{Preliminaries}
\noindent\textbf{OmniGen.}
 OmniGen is built on the Phi-3 framework \cite{abdin2024phi3technicalreporthighly}, which consists of 32 encoder layers and uses Phi-3’s tokenizer to process text without modifications. For image processing, OmniGen employs a VAE to extract latent representations, which are flattened into a sequence of visual tokens with standard frequency-based positional embeddings \cite{Peebles2022DiT}. During inference, OmniGen samples a Gaussian noise $\mathbb{N}$ and applies the flow-matching method \cite{lipman2023flowmatchinggenerativemodeling} to generate the final image.

\noindent\textbf{Attention Mechanism in OmniGen.}
OmniGen applies causal attention to each element in the sequence but applies bidirectional attention within each image sequence. 
The order of the input image and text instruction has minimal impact on the generated image \cite{xiao2024omnigen}. Therefore, in this paper, we focus solely on how the input image and text instruction influence the denoising process, specifically cross-attention, without considering the mutual influence between the image and text.
OmniGen’s attention mechanism does not explicitly include cross-attention. For clarity in the following discussion, we extract the cross-attention component from the joint self-attention as follows:
\begin{equation}
    \mathbb{A}_c  = \{A_{ij} \quad i \in \mathbb{N} ,j\in\mathbb{I}+\mathbb{T}\}
    \label{eq:cross_attn}
\end{equation}
where $A$ is the attention matrix, $\mathbb{N}$ represents the noise latent tokens, $\mathbb{I}$ is the input image tokens and $\mathbb{T}$ represents the text instruction tokens. Unless otherwise specified, the term ``cross-attention" in the following refers to this definition.

\subsection{Vital Steps and Layers}
\begin{figure}[h]
    \centering
    \includegraphics[width=\linewidth]{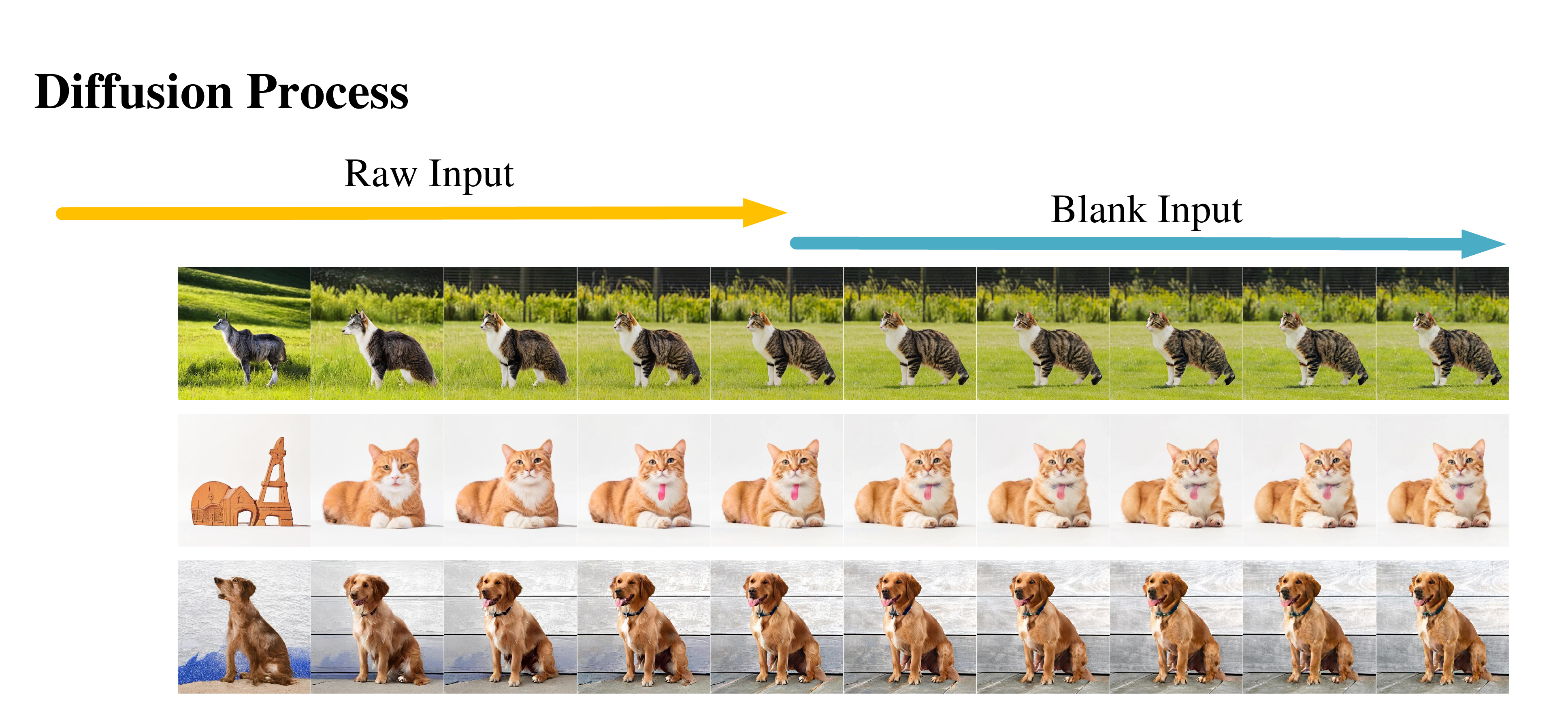}
    \caption{\textbf{Step-wise input perturbation.} From left to right, the images are generated after perturbing the input every 5 steps within the range of 0 to 50 steps.}
    \label{fig:earlystep}
\end{figure}

\noindent\textbf{Step-wise input perturbation.}
We replaced the raw input with a blank one (a pure white image with a blank instruction filled with padding tokens) at different diffusion steps to perturb the denoising process.
As shown in Fig. \ref{fig:earlystep}, perturbations after 20 steps have minimal impact on the generated images, indicating that the input becomes negligible in later stages.  

\begin{figure}[h]
    \centering
    \includegraphics[width=\linewidth]{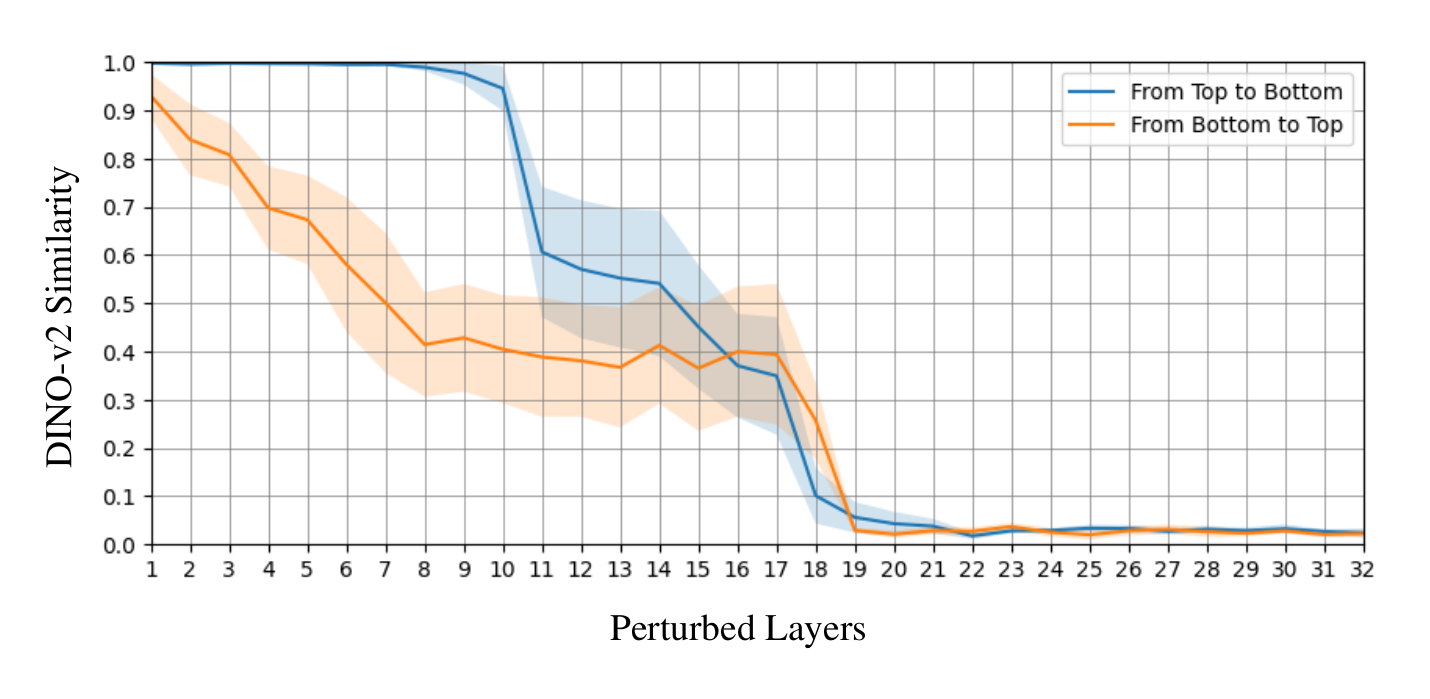}
    \caption{\textbf{DINO-v2 similarity \cite{oquab2023dinov2} by perturbed layers.} Comparing the similarity from top to bottom (blue curve) and from bottom to top (yellow curve), the perturbations in shallow layers have minimal effect on the image generation process.}
    \label{fig:DINOv2 score by disturbed layers}
\end{figure}

\noindent\textbf{Layer-wise input perturbation.}
We extended our input perturbation experiments by replacing layer inputs with blank ones, progressively increasing the number of perturbed layers from top to bottom and vice versa. As shown in Fig. \ref{fig:DINOv2 score by disturbed layers}, perturbing \emph{from bottom to top} leads to a steady drop in DINO-v2 similarity \cite{oquab2023dinov2}, whereas perturbing \emph{from top to bottom} has minimal impact in shallow layers. This suggests that input influence on image generation is negligible in the shallow layers.

\subsection{Self-Adaptive Attention Scaling}

\begin{figure*}[h]
    \centering
    \includegraphics[width=\linewidth]{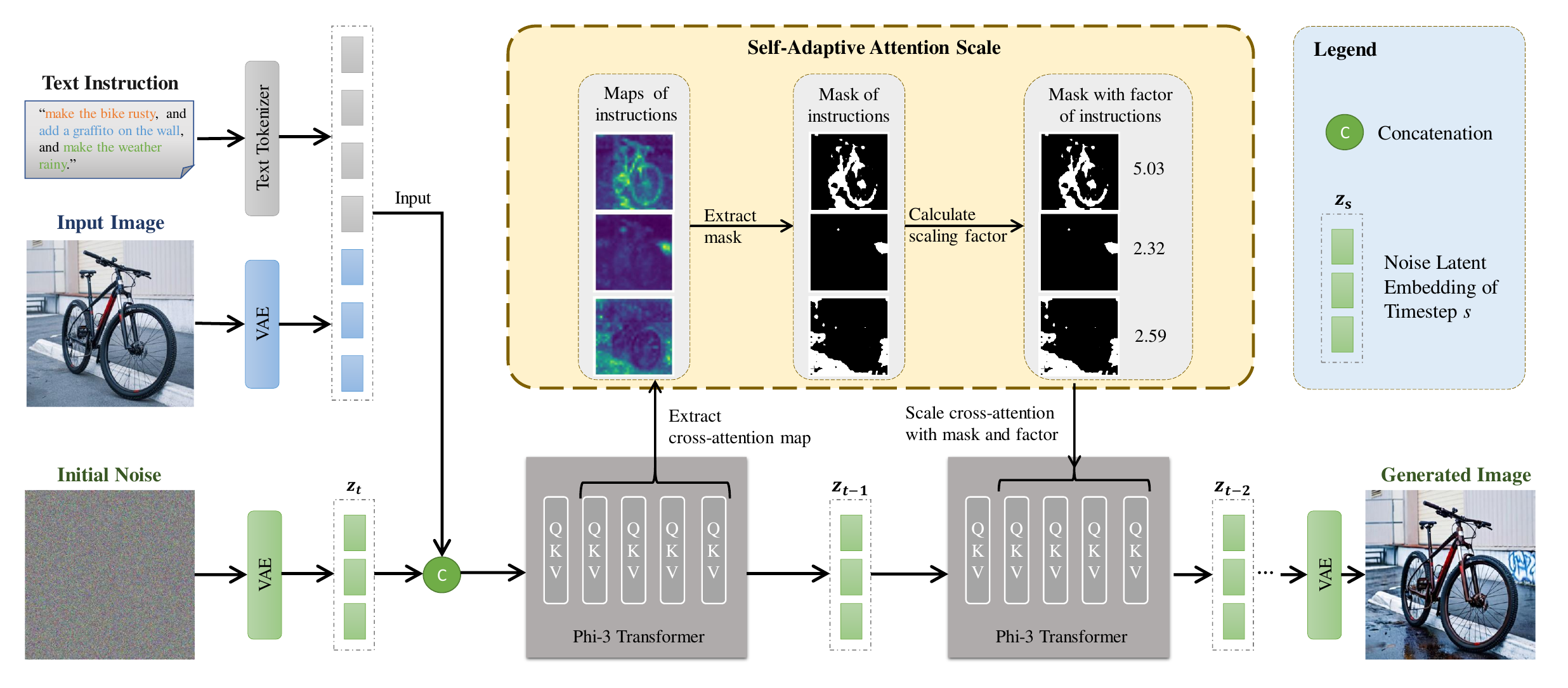}
    \caption{\textbf{Framework of SaaS.} SaaS is designed to enhance the instruction-following fidelity of unified image generation models. At the denoising step $t$, a unique mask of each sub-instruction is extracted according to the cross-attention map averaged by the vital layers. Subsequently, we calculate the ratio of the attention map values of the input image in each masked region to the corresponding sub-instruction's map values, which we term as the scaling factor. At the next denoising step $t-1$, we scale the cross-attention map values within each masked region corresponding to each sub-instruction according to the scaling factor. {With our SaaS, the previously overlooked ``rust'' appears on the bicycle.}}
    \label{fig:Framework}
\end{figure*}

Given the input image $I$ and the composite instruction $T$, consisting of $k$ sub-instructions $\{T_1, T_2,\dots,T_k \}$, our goal is to ensure that all of $\{T_i\}$ are represented in the generated image.
We {assume} that some sub-instructions are overlooked due to the high conflict between their activation regions in the cross-attention maps and the activation region of the input image tokens. To address this, we propose \textbf{SaaS} framework, as illustrated in Fig. \ref{fig:Framework}.  By leveraging the consistency of cross-attention map between adjacent timesteps, at timestep $t$, we extract the masks for each sub-instruction (Sec. \ref{sec:Instruction Masks Extraction}) and calculate the corresponding scaling factors (Sec. \ref{sec:Scale Factor Calculation}). 
At timestep $t-1$, we apply these scaling factors to scale the activation value of the corresponding sub-instruction within the masked regions (Sec. \ref{sec:Cross-attention Scale}).

\subsubsection{Instruction Masks Extraction}
\label{sec:Instruction Masks Extraction}
Inspired by the {reasoning} capabilities of diffusion models \cite{wu2023diffumask,karazija2024diffusionmodelsopenvocabularysegmentation,li2023grounded}, our analysis of OmniGen reveals that we can identify the influence regions on the generated image of the input image and each sub-instruction. For example, as shown in Fig. \ref{fig:crossattn}, we can predict the approximate regions where `graffito' will appear by analyzing the activation of the sub-instruction `add a graffito on the wall' within the cross-attention map. 

Previous analysis has shown that the shallow layers have limited influence on the noise latent $\mathbb{N}$, so we ignore the cross-attention from these shallow layers.  We calculate the average cross-attention across layers and heads and then convert this into a 32x32 cross-attention map $A[e_i]$ corresponding to each token $e_i$ (assuming the generated image resolution is 512x512).

For each sub-instruction $T_i$, at the denoising step $t$, we begin by applying a Gaussian filter \cite{chefer2023attendandexcite,guo2023focusinstructionfinegrainedmultiinstruction} to the corresponding cross-attention map $A_t[e_i] \in \mathbb{R}^{ 32 \times 32}$ to smooth the map. Subsequently, we obtain the map of the entire sub-instruction $T_i$ by summing the maps of all tokens in $T_i$, as described by the following equation:
\begin{equation}
A_t[T_i] = \sum_{e_i\in{T_i}}G(A_t[e_i])    
\end{equation}

where $G$ represents the Gaussian filter.

To extract the mask $M_t[{T_i}]$, we apply a min-max normalization to scale the values in $A_t[T_i]$ to $[0,1]$ range. We then apply a threshold $\tau$ to compute the mask as follows:
\begin{equation}
M_t[T_i] = \mathbbm{1} \left(\text{norm}\left(A_t[T_i]\right)\geq \tau\right) 
\end{equation}

The resulting mask, denoting the region of interest of sub-instruction $T_i$, has dimensions $\in \mathbb{R}^{32\times32}$. Fig. \ref{fig:Framework} shows the mask of each sub-instruction. 

\subsubsection{Scaling Factor Calculation}
\label{sec:Scale Factor Calculation}

In our previous analysis, we identified that some sub-instructions are overlooked due to the conflict between their activation regions in the cross-attention map and those of the input image. As shown in Fig. \ref{fig:crossattn}, both the sub-instruction ``make the bike rusty'' and the input image have activation regions concentrated on the bicycle in the generated image, which leads to the editing instructions being suppressed by the activations of the input image.

To mitigate it, we calculate the ratio $\alpha_t[T_i]$ of the activation values for the input image relative to those of the sub-instruction within the masked region, as detailed as follows: 
\begin{equation}
 \alpha_t[T_i] = \frac{\sum(A_t[I]\cdot M_t[T_i])}{\sum(A_t[T_i]\cdot M_t[T_i])}    
\end{equation}

where $\sum$ denotes the summation of all patches in cross-attention map $A_t \in \mathbb{R}^{32\times 32}$ and $A_t[I]$ represents the map of input image $I$, calculated as $A_t[I] = \sum_{e_i\in I}{G(A_t[e_i])}$.

This ratio serves as the scaling factor, helping to balance the influence of the input image and the sub-instruction.  By adjusting this factor, we ensure that each sub-instruction is appropriately reflected in the generated image. Fig. \ref{fig:Framework} illustrates the scaling factor of each sub-instruction. Notably, scaling factors for neglected instructions are much higher than those for well-processed instructions.

\subsubsection{Dynamic Attention Scaling}
\label{sec:Cross-attention Scale}

Building on the observed consistency of cross-attention maps across adjacent timesteps, we leverage the masks and scaling factors computed at timestep $t$ and apply them to the corresponding sub-instruction tokens at timestep $t-1$.
The {weighted cross-attention maps based on the scaling factors} are formulated as follows: 
\begin{equation}
    A^{'}_{t-1}[e_i] = \xi_t\cdot \alpha_t[T_i]\cdot  A_{t-1}[e_i]\cdot M_t[T_i] \quad \text{if} \quad e_i \in T_i
\end{equation}

Here, the coefficient $\xi_t$ is a timestep-related hyperparameter, and for simplicity, we set it to $\mathbf{1}$. 
After applying the scaling, the new cross-attention maps are passed through the attention mechanism. We then normalize the attention to ensure it follows the standard formulation, i.e., the sum of each column in the attention matrix equals $\mathbf{1}$.

% Since the input has a more significant effect during the initial steps of the denoising process, we apply the SaaS method for the first 20 steps. For the remaining 30 steps, we switch to the standard OmniGen sampling approach.

\section{Experiments}
We conducted experiments on two classic visual processing tasks: instruction-based image editing and visual conditional image generation. In instruction-based image editing, we performed experiments on both single instruction and multiple sub-instruction editing. In visual conditional image generation, we carried out two sub-experiments: image generation from depth map and image generation from segmentation map. We also validated the effectiveness of SaaS through ablation studies.

\subsection{Experimental Settings}
\noindent\textbf{Dataset.} 
For instruction-based image editing, we utilized the EMU-Edit \cite{Sheynin2023EmuEP} dataset, which consists of over five thousand images and features seven distinct editing operations, including background alteration and object addition, for single-instruction tasks. For multiple sub-instruction editing, we randomly selected 200 images from the PIE-Bench \cite{ju2023direct}. For each image, we used Qwen2.5-VL \cite{Qwen2.5-VL} to generate a detailed instruction containing 2-4 sub-instructions, along with the target descriptions. We then performed a manual secondary inspection to verify the quality of the instructions.

For visual conditional image generation, we randomly selected 2,000 examples from the MultiGen-20M \cite{qin2023unicontrol} dataset for generation from the depth map and 2,000 examples from the ADE20K \cite{zhou2019semantic} test dataset for generation from the segmentation map.

\noindent\textbf{Metrics.}
For evaluation, we focus on four primary metrics: CLIP-I \cite{radford2021learningtransferablevisualmodels}, DINO-v2 \cite{oquab2023dinov2}, CLIP-T \cite{radford2021learningtransferablevisualmodels}, and PickScore \cite{Kirstain2023PickaPicAO}.
CLIP-I and DINO-v2 are used to measure image similarity between the generated image and the input image in instruction-based editing tasks, as well as the similarity between the generated image and the ground truth in the dataset for visual conditional image generation tasks.
CLIP-T calculates the text-image similarity between the generated image and the target caption. PickScore is employed to evaluate how well the generated image aligns with human preferences.

\noindent\textbf{Baseline.}
For instruction-based editing, we compare our method with several state-of-the-art (SOTA) instruction-based image editing approaches, including IP2P \cite{brooks2022instructpix2pix}, MagicBrush \cite{Zhang2023MagicBrush}, and OmniGen \cite{xiao2024omnigen}. IP2P serves as the foundational model for instruction-based image editing methods. MagicBrush builds upon IP2P by fine-tuning it on a high-quality, custom dataset. 
For visual conditional image generation tasks, we compare our method with OmniGen.

\begin{figure*}[ht]
    \centering
    \includegraphics[width=\linewidth]{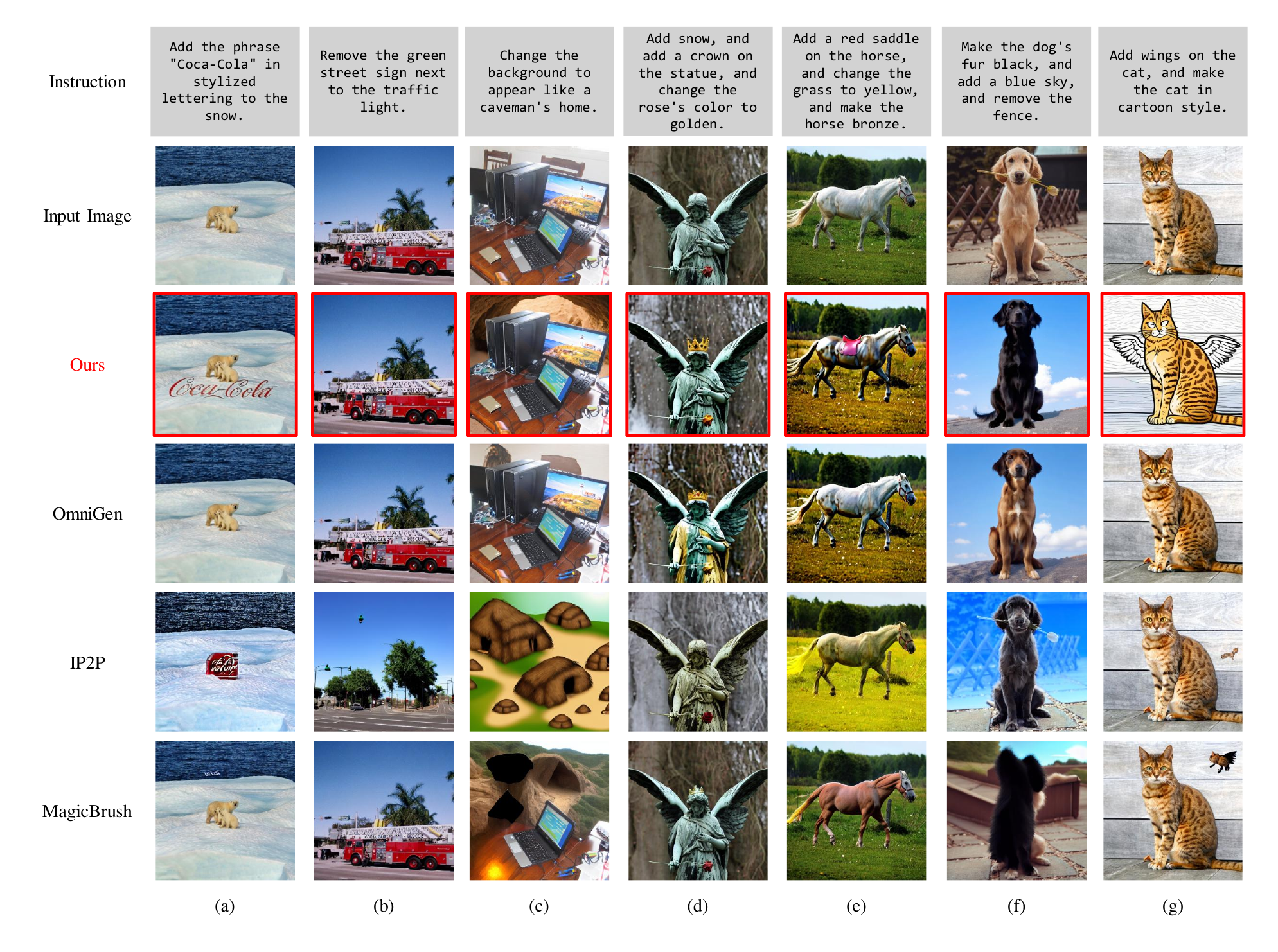}
    \vspace{-1.5em}
    \caption{\textbf{Qualitative comparisons of instruction-based image editing.} We present edited images from the baselines using the same input for each column. From top to bottom: input image, our method, OmniGen \cite{xiao2024omnigen}, IP2P \cite{brooks2022instructpix2pix}, and MagicBrush \cite{Zhang2023MagicBrush}. The text above each column represents the corresponding editing instruction. 
    Compared to these baseline methods, our approach demonstrates superior instruction-following fidelity.
    Zoom in for better visualization.}
    \label{fig:Main_results}
\end{figure*}

\noindent\textbf{Implementation details.}
In all our experiments, we utilize the OmniGen-v1 model with a total of 50 denoising steps. The default OmniGen sampling settings of image guidance $S_I = 1.6$ and text guidance $S_T = 2.5$ are used unless specified otherwise. The resolution of the input image and the generation setting is $512\times512$. For the mask extraction described in Sec. \ref{sec:Instruction Masks Extraction}, in instruction-based image editing, the threshold $\tau = 0.4$, while in visual conditional image generation, the threshold is set to $0.2$. The SaaS is employed for the initial 20 steps, and for the remaining 30 steps, we switch to the standard OmniGen sampling.

\subsection{Instruction-Based Image Editing Results}
For instruction-based image editing, we conducted experiments on two sub-tasks: single instruction editing and multiple sub-instruction editing.

\noindent\textbf{Qualitative Evaluation.}
We present qualitative results in Fig. \ref{fig:Main_results}. From top to bottom, each image shows the input image, the result from our method, OmniGen, IP2P, and MagicBrush, respectively. The text above each column represents the corresponding editing instructions.

The first three columns illustrate results for single-instruction tasks. In object addition (Fig. \ref{fig:Main_results} (a)) and object removal (Fig. \ref{fig:Main_results} (b)), OmniGen and MagicBrush overlook the instruction, while IP2P tends to over-edit. Notably, even for small-scale edits such as ``Remove the green street sign" in Fig. \ref{fig:Main_results} (b), our method adheres closely to the instruction. In background alteration (Fig. \ref{fig:Main_results} (c)), OmniGen only changes the chairs in the background to stones, whereas our method successfully alters the entire background. IP2P and MagicBrush both exhibit some over-editing and lower image quality.

The last four columns show results for tasks involving multiple sub-instructions. The baseline methods neglect some sub-instructions, while our method follows each sub-instruction effectively. For example, in Fig. \ref{fig:Main_results} (d), OmniGen causes region leakage for the sub-instruction ``change the rose's color to golden",  producing a flat yellow color that lacks the specified metallic luster.
Our method executes this sub-instruction correctly, while IP2P and MagicBrush ignore it entirely. In Fig. \ref{fig:Main_results} (e), OmniGen fails to add the saddle; in Fig. \ref{fig:Main_results} (f), OmniGen doesn't turn the dog black as instructed; and in Fig. \ref{fig:Main_results} (g), OmniGen completely overlooks the instruction, with the other two baselines also exhibiting varying levels of instruction neglect.

These qualitative results demonstrate that our method achieves better instruction-following fidelity, leading to superior editing outcomes, particularly in tasks with multiple sub-instructions.
More results of instruction-based image editing are available in the Supplementary Material.

\noindent\textbf{Quantitative Evaluation.}
As illustrated in Tab. \ref{tbl:quantitaive_results}, we compared our method with these baselines in terms of quantitative results. 
Our method achieves state-of-the-art performance on both the CLIP-T and PickScore metrics, demonstrating superior instruction-following fidelity and better alignment with human aesthetics. Notably, for tasks involving multiple sub-instructions, our method shows a larger improvement on the PickScore metric, highlighting its enhanced ability to handle complex multi-instruction editing tasks.

Although our method performs slightly lower than OmniGen on image similarity metrics like CLIP-I and DINO-v2, this is primarily due to OmniGen’s instruction neglect. As a result, some of its output images are very similar to the input images, leading to higher similarity scores. This further emphasizes that our method offers stronger instruction-following fidelity.

\begin{table}[h]
\centering
\resizebox{\linewidth}{!}{
\begin{tabular}{ll c c  c c }
\toprule
\textbf{Edit Task}
& \textbf{Method}
& \shortstack[c]{\textbf{CLIP-I $\uparrow$} } 
& \shortstack[c]{\textbf{DINO-v2 $\uparrow$} }
& \shortstack[c]{\textbf{CLIP-T $\uparrow$}}
& \shortstack[c]{\textbf{PickScore $\uparrow$}}\\
\midrule
\multirow{4}{*}{\shortstack[c]{\textbf{Single}\\ \textbf{Instruction} } }  
& IP2P~\cite{brooks2022instructpix2pix} & 0.810 & 0.613 & 0.244 & 0.146  \\
& MagicBrush~\cite{Zhang2023MagicBrush} & 0.857 & 0.706 & 0.247 & 0.152  \\
& OmniGen~\cite{xiao2024omnigen}   & 0.915 & 0.842 & 0.256 & 0.233 \\
& \textbf{SaaS (ours)}    & 0.900 & 0.835 & \textbf{0.282} & \textbf{0.462} \\
\midrule

\multirow{4}{*}{\shortstack[c]{\textbf{Multiple}\\ \textbf{Sub-instruction} } } 

& IP2P~\cite{brooks2022instructpix2pix} & 0.832 & 0.580 & 0.264 & 0.112 \\
& MagicBrush~\cite{Zhang2023MagicBrush} & 0.863 & 0.751 & 0.261 & 0.131  \\
& OmniGen~\cite{xiao2024omnigen}   & 0.911 & 0.809 & 0.276 & 0.244 \\
& \textbf{SaaS (ours)}     & 0.892 & 0.786 & \textbf{0.315} & \textbf{0.513} \\
\bottomrule
%\vspace{-10pt}
\end{tabular}
}
\vspace{-4pt}
\caption{\textbf{Quantitative comparison of instruction-based image editing.} We compare our method with these baselines in terms of CLIP-I similarity, DINO-v2 similarity, CLIP-T similarity, and PickScore. Our method achieves state-of-the-art results in CLIP-T similarity and PickScore.}
\label{tbl:quantitaive_results}
%\vspace{-6pt}
\end{table}

\noindent\textbf{User Study.}
To better reflect human subjective perception, we conducted a user study with 32 participants, each tasked with selecting the best-edited image.  Each participant evaluated 50 image pairs across both single-instruction and multi-sub-instruction editing tasks. As shown in Tab. \ref{tab:user_study}, the results indicate a strong preference for our method, consistently outperforming alternatives in both task settings.

\begin{table}[h]
\centering
\resizebox{\linewidth}{!}{
\begin{tabular}{l c c c c}
\toprule
 {Edit Task} & IP2P & MagicBrush & OmniGen & SaaS (ours) \\
\midrule
Single Instruction & 10.3\% & 11.6\% & 21.8\% & \textbf{56.3\%} \\
Multiple Sub-instruction & 6.2\% & 7.6\% & 21.0\% & \textbf{65.2\%} \\
\bottomrule
\end{tabular}
}
\vspace{-4pt}
\caption{Results from a user study with 32 participants. Our SaaS outperforms others.}
\label{tab:user_study}
% \vspace{-1.0em}
\end{table}

\begin{figure*}[h]
    \centering
    \includegraphics[width=\linewidth]{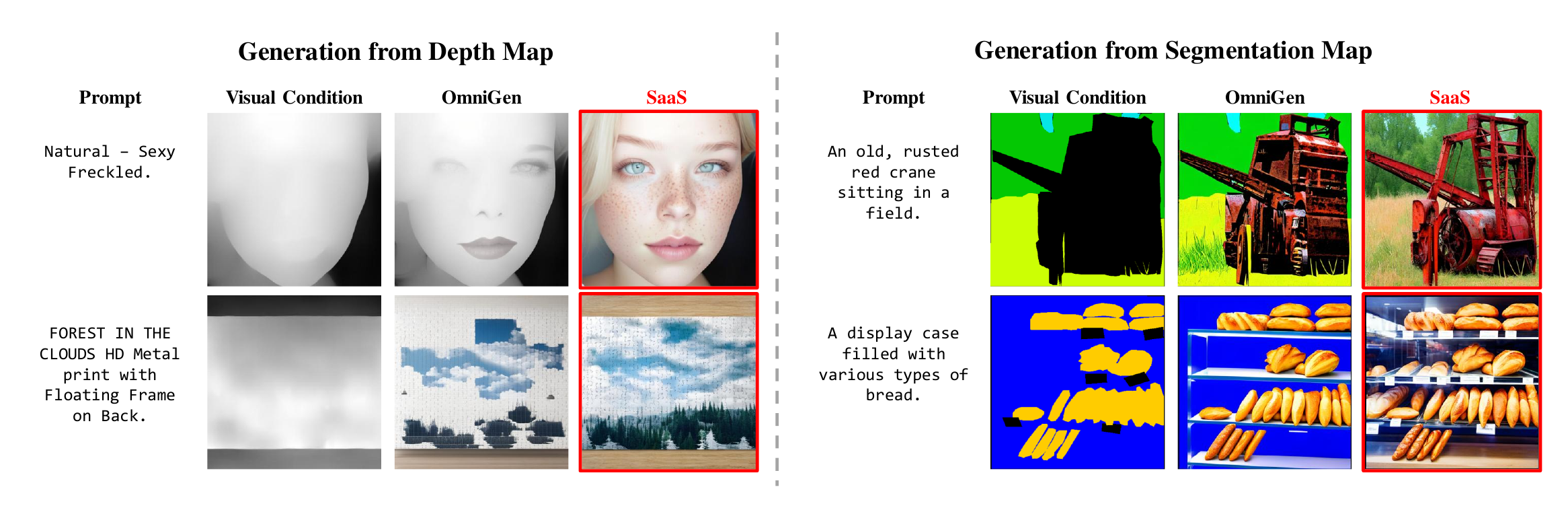}
    \vspace{-1em}
    \caption{\textbf{Qualitative comparison of visual conditional image generation.} The images on the left are generated from the depth map, while those on the right are generated from the segmentation map. Our SaaS method outperforms OmniGen in both instruction-following fidelity and image quality. Zoom in for better visualization.}
    \label{fig:visual_condition}
\end{figure*}

\subsection{Visual Conditional Image Generation Results}
For visual conditional image generation, we conducted experiments on two sub-tasks: image generation from depth map and image generation from segmentation map.

\noindent\textbf{Qualitative Evaluation.}
We present some qualitative results in Fig. \ref{fig:visual_condition}. As seen, OmniGen also exhibits prompt neglect in this visual conditional image generation task. For instance, the freckles are not well generated in the first row of \emph{generation from depth map}.
Additionally, OmniGen suffers from poor generation quality, such as in \emph{generation from segmentation map}, where the background is poorly generated. Our SaaS method not only enhances instruction-following fidelity but also generates higher-quality images than OmniGen in visual conditional image generation.
More results of visual conditional image generation are available in the Supplementary Material.

\noindent\textbf{Quantitative Evaluation.}
Tab. \ref{tbl:quantitaive_results_vcig} presents our quantitative results. The results show that our method outperforms OmniGen on all metrics. Notably, unlike image editing tasks, our CLIP-I and DINO-v2 metrics are calculated by comparing the generated images with the ground truth provided in the dataset. Given the inherent randomness in generation, these two metrics are relatively low. However, even so, our method outperforms OmniGen on them, further indicating that the quality of the images generated by our method is superior.

\begin{table}[h]
\centering
\vspace{-4pt}
\resizebox{\linewidth}{!}{
\begin{tabular}{ll c c  c c }
\toprule
\textbf{Visual Condition}
& \textbf{Method}
& \shortstack[c]{\textbf{CLIP-I $\uparrow$} } 
& \shortstack[c]{\textbf{DINO-v2 $\uparrow$} }
& \shortstack[c]{\textbf{CLIP-T $\uparrow$}}
& \shortstack[c]{\textbf{PickScore $\uparrow$}}\\
\midrule
\multirow{2}{*}{\shortstack[c]{\textbf{Depth}} }  
& OmniGen~\cite{xiao2024omnigen}   & 0.762 & 0.562 & 0.254 & 0.342 \\
& \textbf{SaaS (ours)}    & 0.803 & 0.593 & \textbf{0.296} & \textbf{0.658} \\
\midrule

\multirow{2}{*}{\shortstack[c]{\textbf{Segmentation}} } 
& OmniGen~\cite{xiao2024omnigen}   & 0.721 & 0.549 & 0.274 & 0.317 \\
& \textbf{SaaS (ours)}     & 0.781 & 0.583 & \textbf{0.317} & \textbf{0.683} \\
\bottomrule
%\vspace{-10pt}
\end{tabular}
}
\vspace{-4pt}
\caption{\textbf{Quantitative comparisons of visual conditional image generation.} Our SaaS outperforms OmniGen on all metrics.}
\label{tbl:quantitaive_results_vcig}
\vspace{-1.0em}
\end{table}

\subsection{Ablation Study}

\begin{figure}[h]
    \centering
    % \vspace{-1.5em}
    \includegraphics[width=\linewidth]{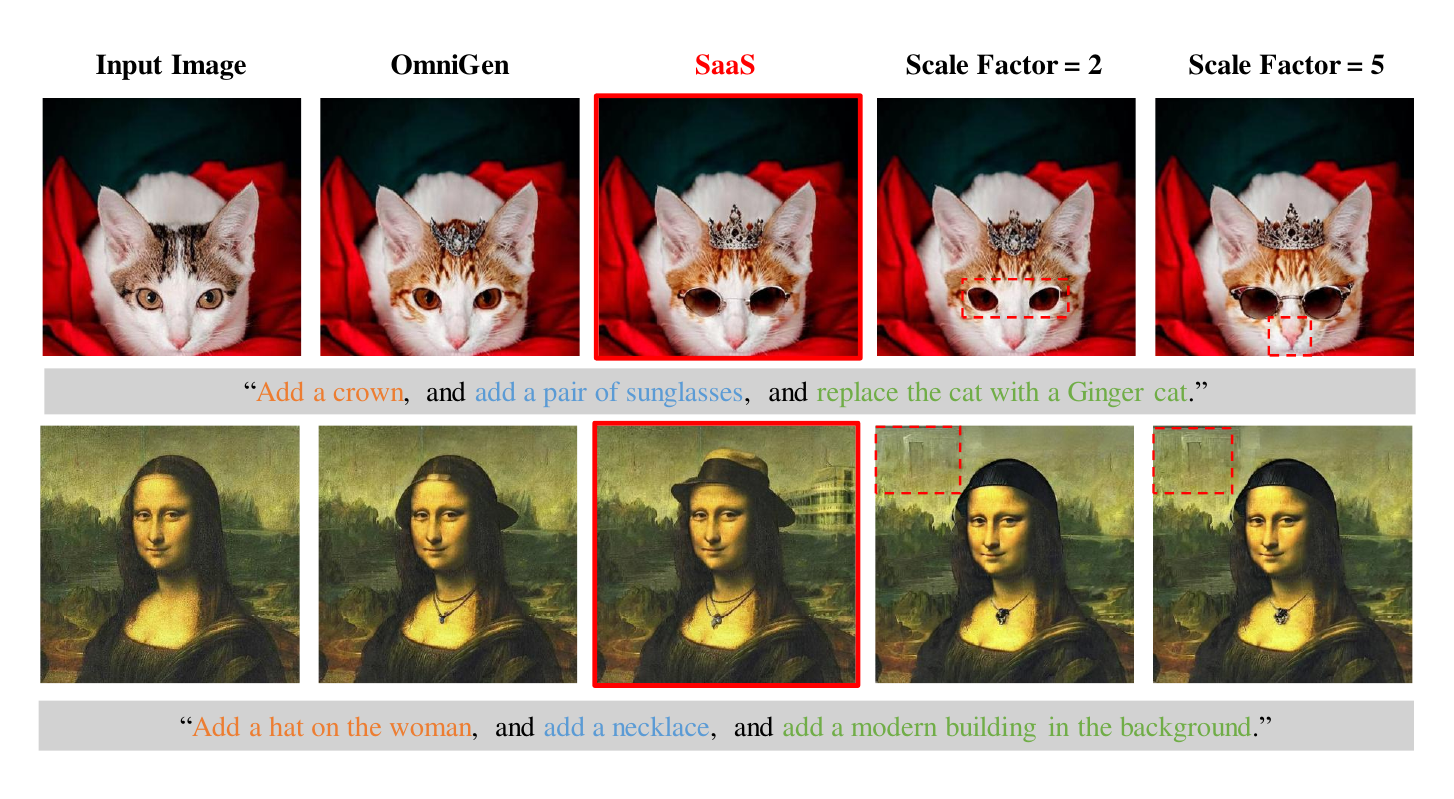}
    \vspace{-1.5em}
    \caption{\textbf{Qualitative comparisons between SaaS and scale with a fixed factor.} We compared SaaS with fixed scaling factors of 2 and 5. The results show that while direct scaling has some effect, our SaaS method outperforms this fixed scaling approach. 
    Zoom in for better visualization.}
    \label{fig:saas_vs_cas}
    % \vspace{-1em}
\end{figure}

\noindent\textbf{Effectiveness of Self-Adaptive Attention Scaling.}
A simpler alternative to our method is applying a fixed scaling factor to cross-attention maps to address instruction neglect. However, this approach has fundamental limitations, as our comparisons in Fig. \ref{fig:saas_vs_cas} demonstrate.
The core issue is that a single, fixed factor is rarely optimal. It may be too weak for some instructions while being too strong for others, creating a trade-off between instruction neglect and over-editing. For instance, a scaling factor of 5 may execute one instruction correctly (Fig. \ref{fig:saas_vs_cas}, top row) but fail on another (bottom row). Similarly, within a single image, a factor sufficient for one sub-instruction (e.g., adding sunglasses) might cause artifacts in other regions (e.g., the nose).
In contrast, our SaaS method dynamically adjusts scaling factors for different sub-instructions. This adaptive strategy ensures high fidelity to complex instructions while maintaining overall visual consistency, effectively resolving the limitations of a fixed approach.

More ablation studies are available in the Supplementary Material.

\section{Conclusion}

In this paper, we focus on the issue of instruction neglect in unified image generation models. Starting from the generation process, we identified vital steps and layers through input perturbation analysis. By analyzing the cross-attention maps at these critical steps and layers, we attributed the instruction neglect problem to significant conflicts between the activated regions of the neglected instructions and the input image in the cross-attention maps. To address this, we propose SaaS, a free lunch to enhance instruction-following fidelity through self-adaptive attention scaling. 
Experimental results on both instruction-based editing and visual conditional image generation validate the effectiveness of our SaaS. 
We hope that our exploration in unified image generation models, along with the proposed method, will inspire future research in related generation and editing tasks.

% \noindent\textbf{Acknowledgment} 
\section*{Acknowledgment}
This work was supported in part by the Natural Science Foundation of China under Grant 62121002.
\cleardoublepage

{
    \small
    \bibliographystyle{ieeenat_fullname}
    \bibliography{main}
}

\clearpage
\begin{appendices}
\section*{Appendices}
\section{Generalizability of SaaS}
\begin{figure}[h]
    \centering
    \includegraphics[width=\linewidth]{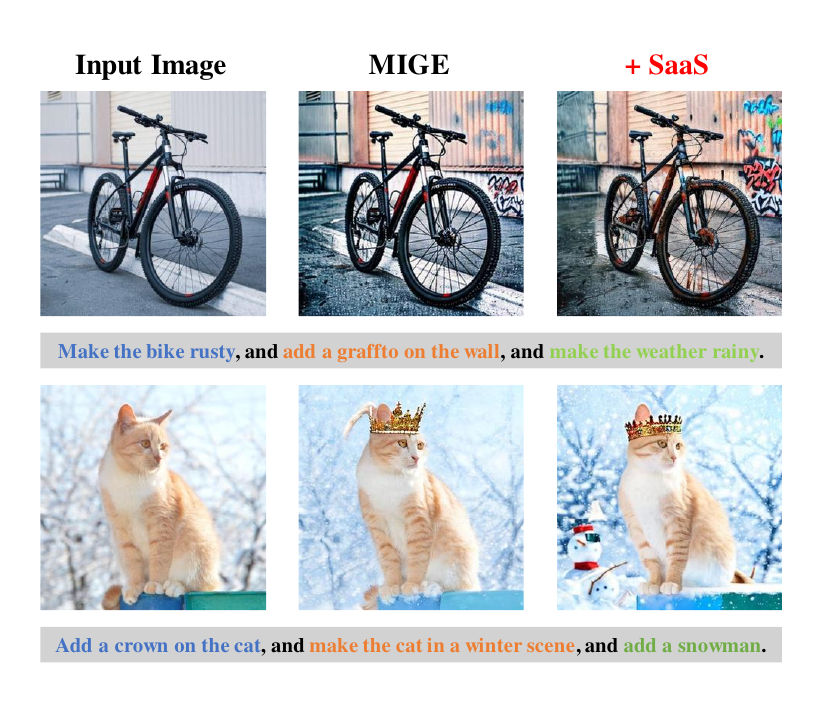}
    \caption{Cases of SaaS on MIGE. Zoom in for better visualization.}
    \label{fig:mige}
\end{figure}

The core principle of our Self-Adaptive Attention Scaling (SaaS) method, adaptively rescaling attention activations between image and instruction tokens, is theoretically model-agnostic, suggesting it should be compatible with various unified image generation architectures. To verify this generalizability, we integrated SaaS into the recently open-sourced MIGE~\cite{tian2025mige} model, a multimodal editing framework distinct from the one used in our main paper.
The results in Fig. \ref{fig:mige} show a significant improvement in instruction following. While the baseline MIGE model struggles with multi-part prompts (e.g., failing to add ``graffiti'' or render a ``snowman''), the SaaS-augmented version successfully executes all sub-tasks. This confirms that SaaS is not over-fitted but serves as a versatile module for enhancing instruction fidelity across different multimodal editing architectures.

\section{Computational Overhead of SaaS}
To verify the practicality of our method, we analyzed the computational overhead introduced by SaaS. We benchmarked inference latency and peak VRAM usage on an NVIDIA RTX A6000 GPU, comparing the baseline OmniGen model with our SaaS-integrated version. As detailed in Tab. \ref{tab:latency_vram}, the findings show that SaaS is remarkably lightweight, adding a mere 0.3 seconds to latency (1.03\% increase) and only 2MB to VRAM consumption (0.02\% increase). This negligible overhead confirms that the significant improvements in instruction-following fidelity are achieved with virtually no additional computational cost, making SaaS a highly efficient and practical solution.
\begin{table}[h]
\centering
% \resizebox{\linewidth}{!}{
\begin{tabular}{ccccc}
\hline
 &  & OmniGen & +SaaS & IEP (\%) \\ \hline
Latency (s) &  & 29.1 & 29.4 & 1.03 \\ 
VRAM (MB) &  & 9988 & 9990 & 0.02 \\ \hline
\end{tabular}
% }
\caption{IEP means Incremental Expense Proportion.}
\label{tab:latency_vram}
\end{table}

\section{Similar Regions Editing}
\begin{figure}[h]
\centering
\includegraphics[width=\linewidth]{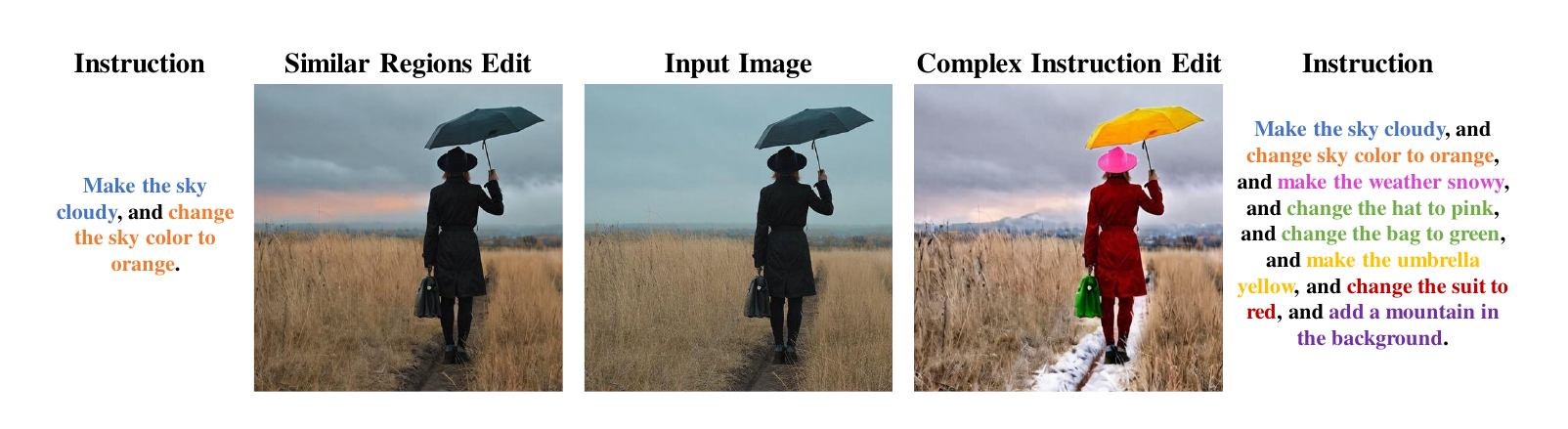}
\caption{Demonstration of SaaS on challenging editing tasks. Left: Accurately editing one of two similar regions. Right: Successfully executing a complex prompt with eight sub-instructions on the same input image.}
\label{fig:similar_eidt}
\end{figure}

Editing visually similar regions is a challenging task requiring precise spatial control. As demonstrated in Figure \ref{fig:similar_eidt}, our SaaS method successfully navigates this challenge. It accurately applies a targeted edit to one of two similar objects (left) and, on the same image, executes a complex prompt with eight sub-instructions (right). This performance highlights SaaS's dual capability in both precise localization and complex instruction following.

\section{Additional Ablation Study}
\begin{figure}[h]
    \centering
    \includegraphics[width=\linewidth]{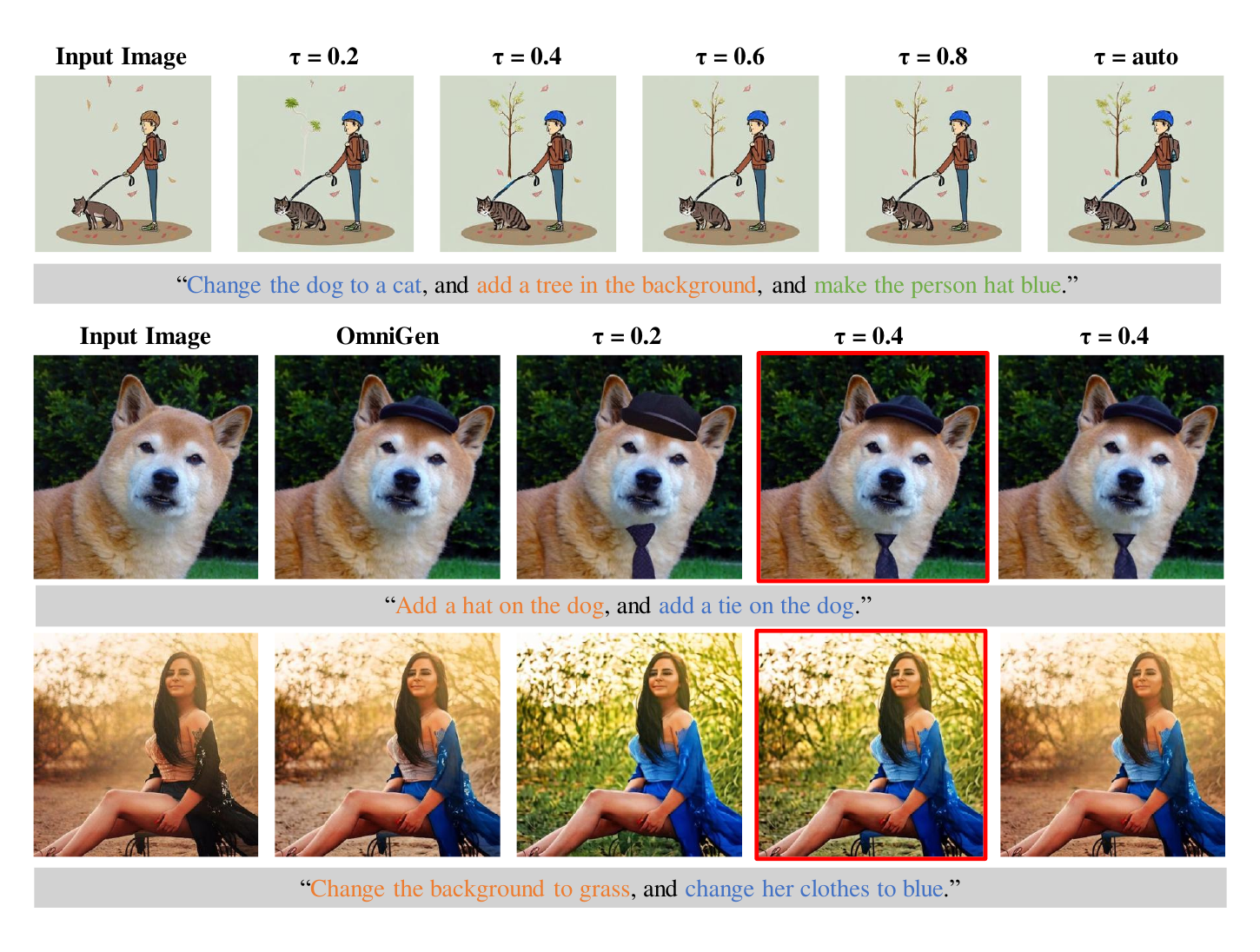}
    \caption{Visual comparison of editing results under different mask thresholds $\tau$. Zoom in for better visualization.}
    \label{fig:auto_mask}
\end{figure}

\noindent\textbf{Mask Threshold.}
In our SaaS framework, the choice of threshold is not critical due to the method's inherent robustness. We provide the Otsu~\cite{xu2011characteristic} method for automatic threshold selection, and as demonstrated in the first row of Fig. \ref{fig:auto_mask} and in Tab. \ref{tab:threshold_pickscore}, different threshold values have minimal impact on the outcome.
Furthermore, as an empirical guideline, lower thresholds work better for global editing, while higher thresholds suit local editing. As illustrated in the second and third rows of Fig. \ref{fig:auto_mask}, a threshold that is too low for local editing can result in an unrealistic appearance, while a threshold that is too high for global editing may cause the edit to fail.

\begin{table}[h]
\centering
\begin{tabular}{cccccc}
\hline
Threshold & 0.2 & 0.4 & 0.6 & 0.8 & auto \\ \hline
PickScore & 0.195 & 0.201 & 0.200 & 0.200 & 0.203 \\ \hline
\end{tabular}
\caption{PickScore values of various thresholds}
\label{tab:threshold_pickscore}
\end{table}

\begin{figure}[h]
    \centering
    \includegraphics[width=\linewidth]{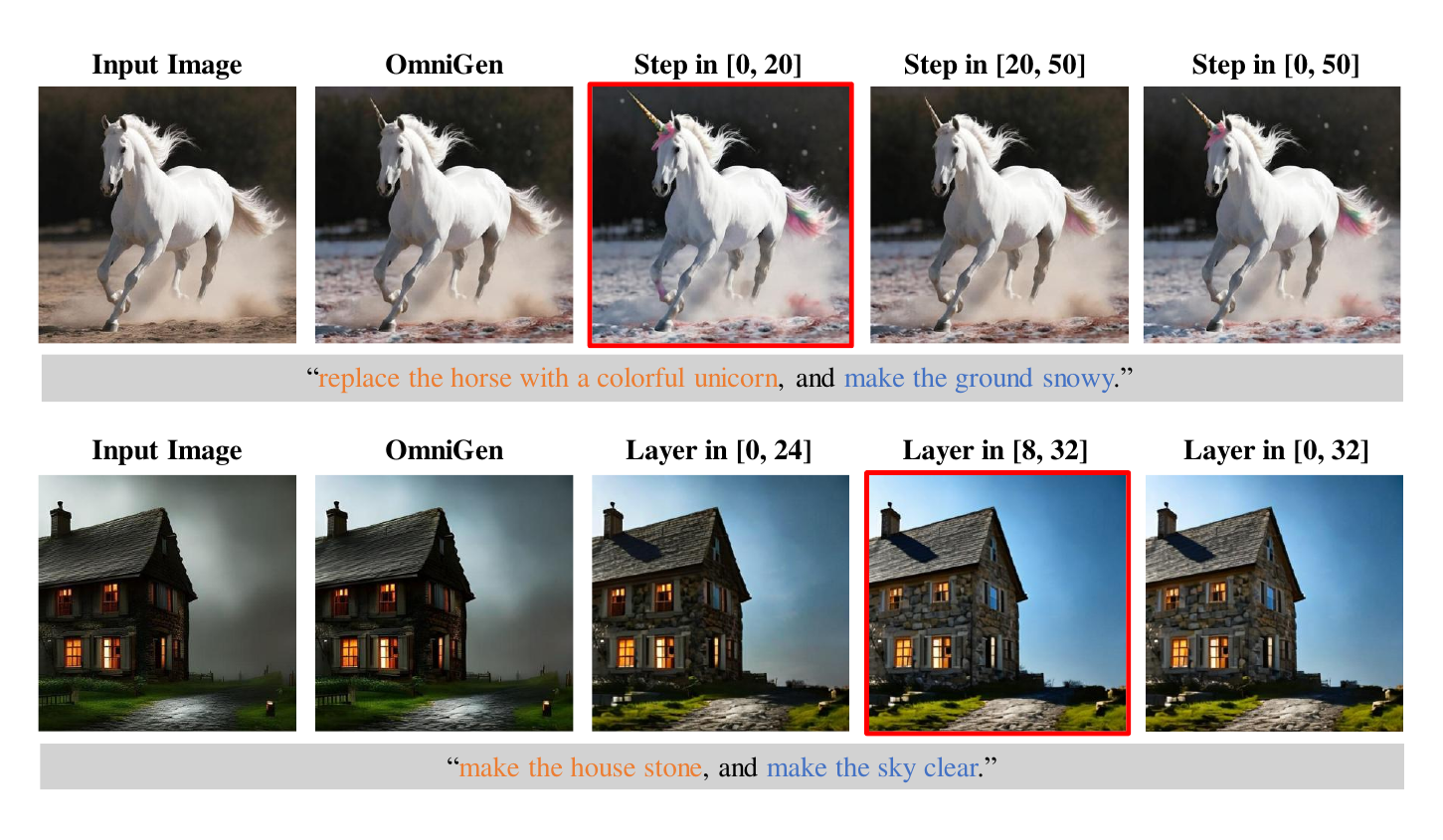}
    \caption{{Visual comparisons between various steps and layers.} Zoom in for better visualization.}
    \label{fig:more_ablation}
\end{figure}

\noindent\textbf{Denoising Steps and Attention Layers.}
Regarding denoising steps, SaaS is more effective when applied in the early stages. As shown in Fig. \ref{fig:more_ablation}, executing SaaS in the early steps achieves similar results to applying it throughout all steps, whereas applying it in the later steps has little to no effect.
Regarding attention layers, SaaS is more effective when applied to deeper layers, yielding results comparable to executing it across all layers. While applying SaaS to shallower layers still has some impact, its effectiveness is noticeably lower than in deeper layers.

\section{Additional Comparison}
\begin{figure*}[ht]
    \centering
    \includegraphics[width=\linewidth]{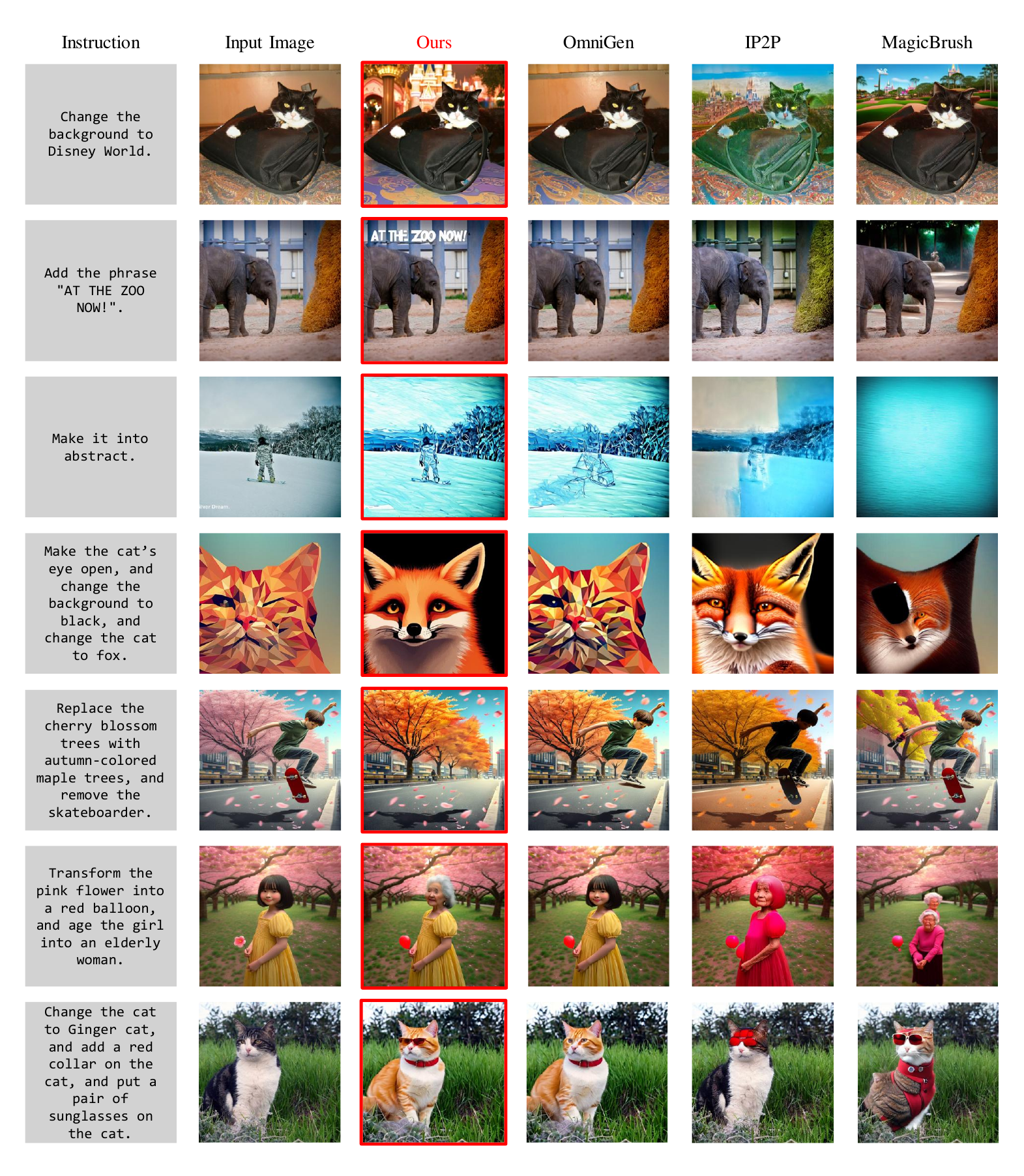}
    \caption{\textbf{Additional qualitative comparison of instruction-based image editing.} We compare our SaaS with these state-of-the-art image editing methods. Zoom in for better visualization.
    }
    \label{fig:more_results}
\end{figure*}

\noindent\textbf{Instruction-based Image Editing.}
In Fig. \ref{fig:more_results}, we provide more qualitative comparison results of our method with other current state-of-the-art methods on the instruction-based image editing task. As can be seen, our method outperforms others in terms of instruction-following fidelity.

Furthermore, we provide a quantitative comparison against several methods: UltraEdit~\cite{zhao2024ultraedit}, ACE++~\cite{mao2025ace++}, and a simple baseline of increasing the guidance scale (Increase Guidance)~\cite{ho2022classifier}. As shown in Tab. \ref{tbl:more_baselines}, our method outperforms these approaches, achieving state-of-the-art (SOTA) results on metrics including CLIP-T and PickScore.

\begin{table}[h]
\centering
\resizebox{\linewidth}{!}{
\begin{tabular}{ll c c  c c }
\toprule
\textbf{Edit Task}
& \textbf{Method}
& \shortstack[c]{\textbf{CLIP-I $\uparrow$} } 
& \shortstack[c]{\textbf{DINO-v2 $\uparrow$} }
& \shortstack[c]{\textbf{CLIP-T $\uparrow$}}
& \shortstack[c]{\textbf{PickScore $\uparrow$}}\\
\midrule
\multirow{4}{*}{\shortstack[c]{\textbf{Single}\\ \textbf{Instruction} } }  
& UltraEdit & 0.876 & 0.750 & 0.266 & 0.228  \\
& ACE++ & 0.941 & 0.855 & 0.249 & 0.152  \\
& Increase Guidance   & 0.879 & 0.732 & 0.262 & 0.228 \\
& \textbf{SaaS (ours)}    & 0.900 & 0.835 & \textbf{0.282} & \textbf{0.397} \\
\midrule

\multirow{4}{*}{\shortstack[c]{\textbf{Multiple}\\ \textbf{Sub-instruction} } } 

& UltraEdit & 0.835 & 0.552 & 0.284 & 0.197 \\
& ACE++ & 0.950 & 0.860 & 0.240 & 0.150  \\
& Increase Guidance   & 0.862 & 0.740 & 0.282 & 0.181 \\
& \textbf{SaaS (ours)}     & 0.892 & 0.786 & \textbf{0.315} & \textbf{0.469} \\
\bottomrule
%\vspace{-10pt}
\end{tabular}
}

\caption{Quantitative comparison on more baselines.}
\label{tbl:more_baselines}

\end{table}

\begin{figure*}[ht]
    \centering
    \includegraphics[width=\linewidth]{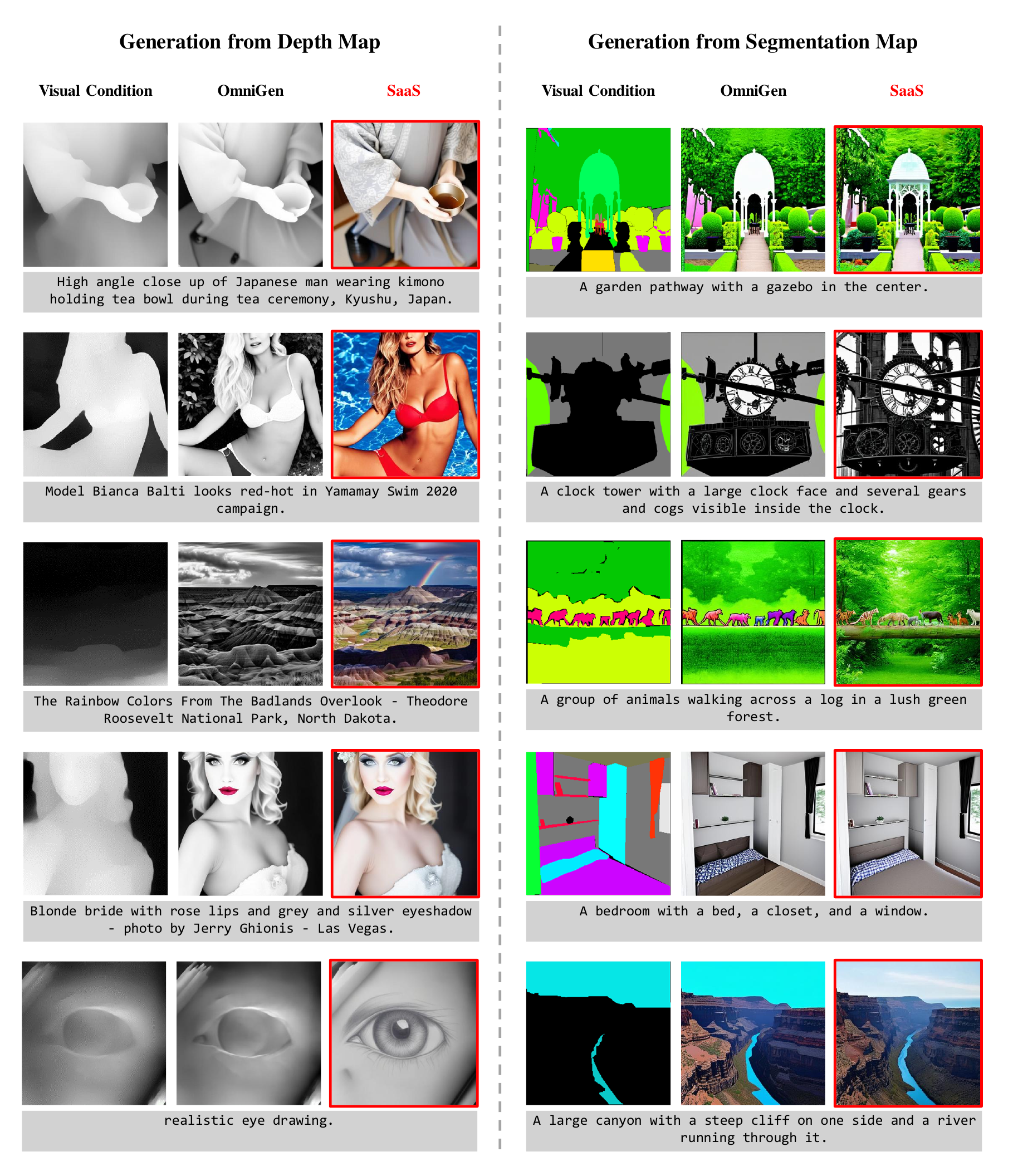}
    \caption{\textbf{Additional qualitative comparison of visual conditional image generation.}
We compare our SaaS method with OmniGen in the generation tasks from the depth map and the segmentation map. The text below each image represents the corresponding instruction. For better visualization, please zoom in.
    }
    \label{fig:more_results_of_vcig}
\end{figure*}

\noindent\textbf{Visual Conditional Image Generation.}
We provide more qualitative results of visual conditional image generation in Fig. \ref{fig:more_results_of_vcig}. 
On the left are images generated from the depth map, and on the right are images generated from the segmentation map. The text below each set of images corresponds to the respective instructions. As can be seen, whether generated from the depth map or segmentation map, our SaaS method demonstrates better instruction-following fidelity and also produces higher-quality images.

\end{appendices}

\end{document}